\pdfoutput=1

\documentclass[11pt]{article}

\usepackage{acl}

\usepackage{times}
\usepackage{latexsym}
\usepackage{amssymb}
\usepackage{graphicx}
\usepackage{url}
\usepackage{booktabs,arydshln}
\usepackage{amsmath,amssymb}

\usepackage{mathtools}
\mathtoolsset{showonlyrefs=true}
\usepackage{multirow}
\usepackage[T1]{fontenc}

\usepackage[utf8]{inputenc}

\usepackage{microtype}
\usepackage{tcolorbox}

\title{Evaluation of Instruction-Following Ability \\for Large Language Models on Story-Ending Generation}

\author{Rem Hida, Junki Ohmura, Toshiyuki Sekiya \\
        Sony Group Corporation, Tokyo, Japan\\
        \{remu.hida, junki.ohmura, toshiyuki.sekiya\}@sony.com }

\begin{document}
\maketitle

\begin{abstract}
Instruction-tuned Large Language Models~(LLMs) have achieved remarkable performance across various benchmark tasks.
While providing instructions to LLMs for guiding their generations is user-friendly, assessing their instruction-following capabilities is still unclarified due to a lack of evaluation metrics.
In this paper, we focus on evaluating the instruction-following ability of LLMs in the context of story-ending generation, which requires diverse and context-specific instructions.
We propose an automatic evaluation pipeline that utilizes a machine reading comprehension (MRC) model to determine whether the generated story-ending reflects instruction.
Our findings demonstrate that our proposed metric aligns with human evaluation.
Furthermore, our experiments confirm that recent open-source LLMs can achieve instruction-following performance close to GPT-3.5, as assessed through automatic evaluation. 
\end{abstract}

\section{Introduction}

\begin{figure}[t]
    \centering
    \includegraphics[width=\linewidth]{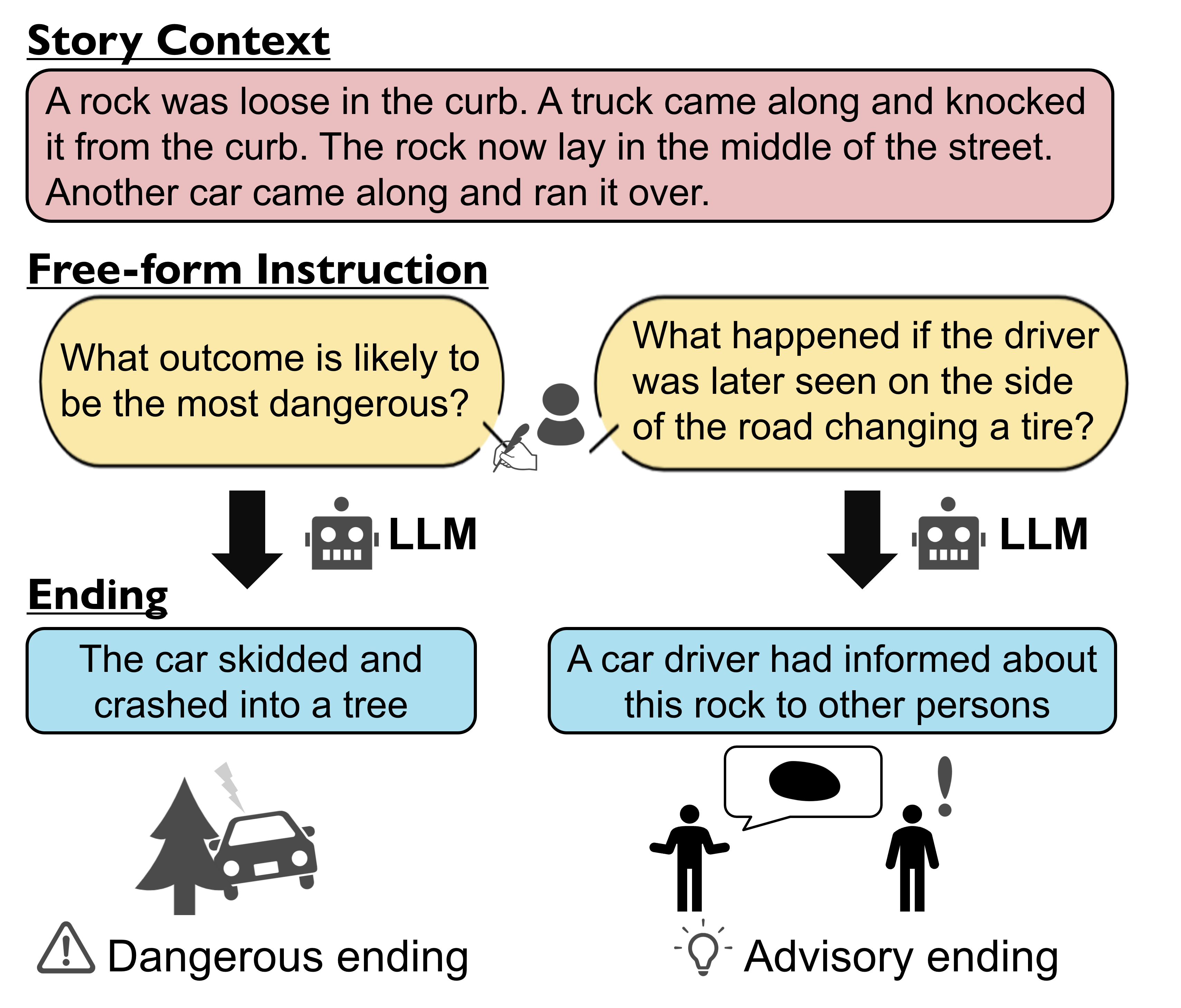}
    \caption{\textbf{Overview of instruction-following story-ending generation:} Conditioning by instruction texts produces different endings such as dangerous and advisory endings. E.g, the examples are from Possible Stories~\cite{ashida-sugawara-2022-possible}.}
    \label{fig:model_image}
\end{figure}

Large Language Models~(LLMs) have already achieved high performance in various NLP tasks~\citep{brown2020language,ouyang2022training,touvron2023llama,anil2023palm}.
One key technique is instruction tuning, which enables LLMs to interpret user inputs even in zero-shot settings on unseen tasks~\citep{chung2022scaling,51119,sanh2022multitask}.
The instruction-following ability of LLMs enables users to utilize LLMs more easily to solve tasks and add constraints, including some keywords in outputs~\citep{pmlr-v202-zhou23g} and answering based on a specific persona~\citep{shanahan2023roleplay}.
The LLMs have been employed in creative writing tasks such as story generation via writing assistant tools~\citep{wordcraft22,ippolito2022creative}.
In story generation, users, who want to write stories, use LLMs to elicit a wide variety of ideas, thus the instruction-following ability is crucial.

Nonetheless, the instruction-following ability is still under-explored; it has been verified in limited tasks and instruction styles.
Regarding tasks, instruction-following ability in summarization has been explored~\citep{skopek-etal-2023-towards, liu2023benchmarking}.
Concerning instruction styles, \citet{zhou2023instructionfollowing} demonstrated the instruction-following ability on verifiable instructions, which are determinable constraints with a binary value to determine if they are being followed or not, such as including keywords, limiting word count, and outputting detectable format like JSON. 
However, LLMs can be used more widely for other tasks and styles of instruction. 

In this paper, we limit our scope to the story-ending generation task via instructions (Figure~\ref{fig:model_image}) and propose a pipeline to evaluate LLMs' instruction-following ability on this task.
The possible instructions in the story-ending generation are more diverse and context-specific than in traditional tasks, such as summarization, or verifiable instructions.
Our evaluation pipeline is twofold: 
i) \textbf{Conversion to Evaluation Dataset}: LLMs generate endings from their input comprising story context and instruction, converted from the existing dataset Possible Stories~\citep{ashida-sugawara-2022-possible}, to create the evaluation dataset.
ii) \textbf{Evaluation via the MRC model}: the MRC model receives the story context, instruction, and candidate endings, which include the generated ending. Then, the model identifies whether the ending follows the instruction based on the context. 
We define overall accuracy as the MRC-based score. 
We confirm that our metric aligns with human evaluation.
Moreover, using this pipeline, we verify that recent open-source LLMs can perform close to GPT-3.5.

\begin{figure*}[t]
    \centering
    \includegraphics[width=\linewidth]{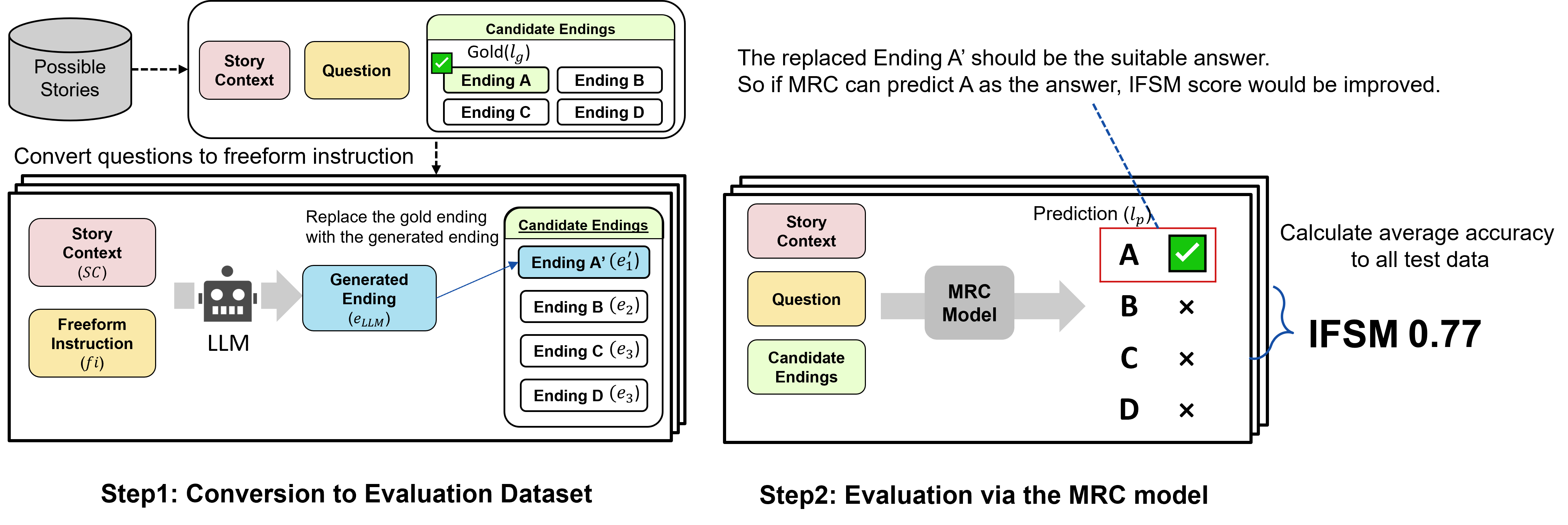}
    \caption{Evaluation pipeline of IFSM~(\textbf{I}nstruction \textbf{F}ollowing \textbf{S}core from the \textbf{M}RC model)}
    \label{fig:eval}
\end{figure*}

\section{Related Work}
\paragraph{Instruction-Following Ability}
While the instruction-following ability remains vague, evaluations have been attempted such as task-specific instruction~\citep{skopek-etal-2023-towards,liu2023benchmarking}, verifiable instruction which can be mechanically determined whether they have achieved like keyword or length constraints~\citep{zhou2023instructionfollowing}.
Furthermore, evaluations using LLMs have been adopted ~\citep{jiang2023followbench,xia2024fofo,zeng2024evaluating,zheng2023judging}.
Unlike the previous work, we demonstrate the ability to follow more diverse instructions on specific tasks.

\paragraph{Application of Instruction-Following Ability}
User interface research using LLMs for writing assistants on story generation has become popular~\citep{CHI22-coauthor,chung2022talebrush, osone2021buncho, MORI2023101484,programmer_iui2022}.
Some work utilizes the instruction-following ability of LLMs to enhance tools functions~\citep{wordcraft22, reza2024abscribe}.
For example, ~\citet{wordcraft22} shows that writing assistant users create diverse instructions for LLMs on story generation tasks.
Unlike previous work, we focus on quantifying LLMs' instruction-following ability, which can contribute to user interface studies.

\paragraph{Conditional Story Generation}
With the development of LLMs, numerous approaches have been proposed for story generation under certain conditions including persona~\citep{zhang-etal-2022-persona}, genre~\citep{10.1145/3485447.3512004, peng2018towards}, style~\citep{reif-etal-2022-recipe}. Moreover, \citet{sun-etal-2023-evaluating} illustrates the prefix following ability in story generation using instructions such as \textit{“Please continue writing this story within 4 very short sentences: <prefix>}. 
Different from previous work, we mainly focus on more diverse conditioning on story generation as free-form instructions.

\section{Instruction-Following Ability on Story-Ending Generation}
\subsection{Task Setting} \label{ssec:task_setting}
In the story-ending generation task guided by instructions, the goal is to generate a story-ending that aligns with the given story context and reflects the provided instruction.
More formally, let $SC=\{sc_{1},...,sc_{n}\}$ be the story context where $sc_{i}$ denotes the $i^{th}$ sentence of the story context, and $fi$ represents the free-from instruction.
Given $SC$ and $fi$ as inputs, the objective is to generate the following story-ending $e_{LLM}$, which can be formulated as $e_{LLM} = \texttt{LLM}(SC,fi)  \label{eq:story_ending_generation}.$

\subsection{Dataset}\label{ssec:dataset}
We used the Possible Stories dataset~\cite{ashida-sugawara-2022-possible} for story-ending generation experiments.
This dataset is derived from ROCStories~\cite{mostafazadeh-etal-2016-corpus} and proposed for MRC tasks.
Possible Stories comprises triples, each containing four sentences of story context, four one-sentence ending candidates, and questions designed to distinguish one ending from the others. 

In our settings~(Figure~\ref{fig:model_image}), we regarded the story context as the input and the ending as the output. 
Additionally, we converted the question~($q$) into a free-form instruction~($fi$) as an additional input using the prompt template, because the questions contain information to generate specific endings, such as dangerous outcomes.
Each context has multiple questions, leading to making diversity.

\subsection{Proposed Metric for Instruction-Following}\label{ssec:metrics}
It is not easy to assess how well the LLMs follow the instructions on the story-ending generation directly due to its diversity and style.
Therefore, we use the MRC model as a proxy metric for instruction-following ability in story-ending generation.
We denote this metric as IFSM~(\textbf{I}nstruction \textbf{F}ollowing \textbf{S}core using the \textbf{M}RC model). 
In this context, the MRC task is formulated as a multiple-choice that selects suitable story-endings aligned with context and questions from the ending options.
This can be formulated as Eq. \eqref{eq:mrc_pred} 
\setlength{\abovedisplayskip}{1pt}
\setlength{\belowdisplayskip}{1pt}

\begin{align}
    l_p = \texttt{MRC}_\texttt{model}(SC,q,(e_{1},e_{2},e_{3},e_{4})), \label{eq:mrc_pred}
\end{align}
where $l_p$ denotes a predicted label from the MRC model and $e_{j}$ represents the $j$th ending candidates.
As for the MRC model, we train the model using half of the original Possible Stories dataset for each train, valid, test split\footnote{The remaining dataset is reserved exclusively for test dataset in the story-ending generations. The details are described in Appendix~\ref{appendix:mrc_model}.}. 
We use the DeBERTa~\citep{he2021deberta}\footnote{We use the DeBERTa-v3 from~\url{https://huggingface.co/microsoft/deberta-v3-base}.} for training because it is the best accuracy model on the original paper with Possible Stories.

Given the MRC model, we obtain IFSM in two steps~(Figure~\ref{fig:eval}). 
\textbf{Step1. Conversion to Evaluation Dataset}: Generate a story-ending based on story context and free-form instructions, which is converted from the question, from instruction-tuned LLMs, then replace each gold answer with the generated ending on the test set of Possible Stories.
For example, if $e_1$ is the gold ending, then we replace it with $e_{LLM}$.
\textbf{Step2. Evaluation via the MRC Model}: Predict a suitable ending based on the question, which serves as the source of instruction, from the endings that contain the replaced candidate. Then check whether the predicted ending label matches the gold label, and calculate the average scores for the overall test set, as in Eq. \eqref{eq:ifsm}.
\begin{align}
    \texttt{IFSM} = \frac{1}{n}\sum_{i=1}^{n}\mathbb{I}[l_p^i=l_g^i], \label{eq:ifsm}
\end{align}
where $l_g$ is the gold label, $i$ denotes the $i$th instance, $n$ represents the number of test dataset, and $\mathbb{I}[a]$ returns 1 if $a$ is true and 0, if otherwise.
Suppose the story generation model can generate a more suitable ending that follows the instructions. 
In that case, a predicted label should match the replaced ending's label, as other endings are created as negative examples for instructions.

\section{Evaluation Experiments}

\subsection{Model}
We used the instruction-tuned LLMs for our comparison, including FLAN-T5(small to xxl)~\citep{chung2022scaling}, Llama2-7B-chat~\citep{touvron2023llama}, Llama3-8B-Instruct~\citep{llama3modelcard}, Mistral-7B~\citep{jiang2023mistral} as open-source LLMs and GPT-3.5 as proprietary LLMs (See Table \ref{table:comparison_models} in Appendix~\ref{comparison_model_info} for the detail).

\subsection{Generation Setting}
The following prompt is used for a generation:
\begin{tcolorbox}[fontupper={\ttfamily \small}, title=Prompt Template]
  Please write an ending in one sentence that follows the Context and is a candidate for the answer to the Question. Write with at most 10 words.
  \\\\
  Context: <context>
  
  Question: <question>
  
  Ending:
\end{tcolorbox}
\noindent
Here, \texttt{<context>} and \texttt{<question>} are from the Possible Stories.
We incorporated a length constraint because some LLMs tend to generate longer endings without length constraints, which can affect a fair comparison. Longer endings can contain more easily conditioned information\footnote{The length constraint detail is described in Appendix~\ref{appendix:length_constraint}.}.

For simplicity, we used default parameters to generate endings for both the open-source and proprietary LLMs.

\subsection{Human Evaluation}\label{sec:human_eval}
We confirmed whether IFSM accurately measured the instruction-following ability of LLMs by comparing human judgment.
To address this, we sampled LLM-generated endings and asked annotators to rate them.
We focused on comparing four LLMs~(FLAN-T5-xxl, Mistral-7B, Llama-3-8B, GPT-3.5) and the oracle for simplicity, and sampled 45 instances.
To balance the instance distribution, we sampled instances that are predicted as gold labels (Follow) and labels other than gold (Not Follow) by the MRC model\footnote{We used 25 and 20 instances as the Follow samples and Not Follow samples, respectively.}.
In our human evaluation setting, annotators were asked to rate the 1-5 scale concerning fluency, coherence, and instruction-following, a higher value indicates a better score.
Before the evaluation, we verified the validity of the wording and the operability of the interface for the human assessment using an annotator who is a business-level English speaker. 
We asked two other internal annotators who are native English speakers to rate these instances. 
Appendix ~\ref{sec:evaluation_ui} describes the evaluation format and interface.

\begin{table}[t!]
    \centering
    \small
    
    \begin{tabular}{@{}llll@{}}
    \toprule
    \multirow{2}{*}{MRC prediction}             & \multicolumn{3}{c}{Human Evaluation Perspectives} \\
     & Fluency & Coherence & \begin{tabular}{l}
     Instruction\\Following\end{tabular} \\ \midrule
    Follow   & 4.50     & 4.12       & 4.10       \\
  Not Follow & 4.55    & 4.10       & 3.05      \\\hdashline
  $\Delta$ & 0.05 &0.02 & \textbf{1.05} \\ \bottomrule
    \end{tabular}
    \caption{\textbf{Human Evaluation Result}: 
    The instruction-following rate has the gap between each label predicted by the MRC model. This indicates that the IFSM aligns with human evaluation.}
    \label{table:human_eval}
\end{table}

Table~\ref{table:human_eval} shows the results of the human evaluation.
A score gap~($\Delta$) exists in the instruction-following perspective between the labels predicted by the MRC model, whereas not much difference exists between the other two perspectives.
This demonstrated the MRC model's prediction is valid for our use-case.
While the score of "Not Follow" in the instruction-following perspective is not low in absolute value, it is crucial to distinguish between "Follow" and "Not Follow" in a relative manner.
This finding indicates that the IFSM can evaluate the instruction-following ability in our settings\footnote{We compared to other automatic evaluations in Appendix~\ref{appendix:human_eval_comaprison}.}.
We also investigated the correlation between evaluations by the two annotators. 
The Pearson's correlations for fluency, coherence, and instruction-following are 0.43, 0.19, and 0.36, respectively.
Despite the subjectivity of the evaluation, it correlates to some extent\footnote{A evaluator notes that coherence is the most difficult because it requires considering various factors.}.

\begin{table}[t!]
\centering
\small
\begin{tabular}{@{}lll@{}}
\toprule
model                    &  IFSM↑ & Dissimilarity↑  \\ \midrule
FLAN-T5-small                 & 0.399      & 0.427              \\
FLAN-T5-base                  & 0.576      & 0.564            \\
FLAN-T5-large                 & 0.660      & 0.608              \\
FLAN-T5-xl                    & 0.684      & 0.568              \\
FLAN-T5-xxl                      & 0.753            &   0.596              \\
Mistral-7B                      & 0.891      & 0.538              \\
Llama2-7B            & 0.891      & 0.603              \\
Llama3-8B       &0.867      & 0.669               \\
GPT-3.5        & 0.858      & 0.629              \\ \midrule
Oracle Endings &   0.868         & 0.720          \\ \bottomrule   
\end{tabular}
    \caption{\textbf{Automatic Evaluation Result}: The recent LLMs has achieve around oracle scores.}
    \label{fig:result_ifa}
\end{table}

\subsection{Automatic Evaluation}\label{sec:automatic_eval}
We confirmed the reliability of IFSM and then applied it to additional data and models for automatic evaluation. 
We introduced a controllability metric (Dissimilarity) other than IFSM because LLMs should generate different endings based on different instructions and IFSM can't evaluate it.
We used LaBSE~\citep{feng-etal-2022-language} to measure the semantic dissimilarity between ending pairs ($e_{i}$, $e_{j}$) from the same context with different instructions, formulated as 1 - $cos(\text{LaBSE}(e_{i}),\text{LaBSE}(e_{j}))$\footnote{We've confirmed using another sentence similarity model.}.
We computed the average dissimilarity for each ending pair from the same context.

Table~\ref{fig:result_ifa} shows the results for each score of the compared models.
Mistral-7B and Llama2-7B perform best on the IFSM, and Llama3-8B achieve high Dissimilarity. 
The open-source LLMs are comparable to GPT-3.5
Although the IFSM of recent LLMs are around the oracle endings created by humans, their dissimilarities are a little behind the oracle.
Further analysis is required in future work.\footnote{The generated examples are shown in Table~\ref{table:examples}.}. 

\section{Conclusion}
In this paper, we evaluated instruction-following ability in story-ending generation task, which requires creativity and elicits various user instructions for LLMs.
We proposed the new metric based on the MRC model to assess instruction-following ability. 
We revealed that our proposed metric aligns with human evaluation and that larger models can achieve high instruction-following ability.
We believe our findings will help researchers quantify the instruction-following capabilities of LLMs beyond easily verifiable instructions.

\section*{Limitation}
This paper demonstrated the instruction-following ability of LLMs on story-ending generation tasks.
However, some limitations should be addressed in future work.
Our investigation is limited by the use of Possible Datasets. 
While we posit that question-style instruction is effective to some extent for story generation, some work have used the instructive-style one.
Moreover, our demonstration is limited to short story-ending generation because the Possible Story is based on ROCStories that contain five sentences.
While we investigated LLMs' instruction-following ability on story-ending generation, this may not reflect the overall performance in instruction-following ability.

Our evaluation was also limited to language. While various multilingual datasets have been proposed, we did not address the multilingual aspect of instruction-following. 
However, our approach converts the existing dataset into an instruction-following evaluation method. 
Thus, it is a promising way to adapt existing multilingual datasets for instruction-following evaluation.

\section*{Ethical Statement}
Potential risks could arise from story contexts and instructions. 
LLMs may follow instructions aimed at eliciting toxic or other harmful information regarding the story generation.
Furthermore, owing to the nature of the task of creating fiction, a significant room may exist for bias to enter, from user inputs and LLMs' generation.
We also conducted a human evaluation of the story-ending generated using LLMs.
In our experiments, we used a few instances to confirm the safety of the generated text in advance.
However, if we scale up our human evaluation and reduce it to assess the safety of the generated texts, they may make annotators uncomfortable due to harmful content.

\bibliography{acl}

\appendix

\section{MRC Model Setting}\label{appendix:mrc_model}
We used the original Possible Stories datasets for MRC model training.
We split the train, validation, and test dataset into two subsets: one for MRC model training and the other for evaluating instruction-following ability in story-ending generation.
The split sizes are described in Table \ref{table:dataset_size}.
We used the Possible Stories datasets repository code\footnote{\url{https://github.com/nii-cl/possible-stories/tree/master}} based on the transformers library for training with some modifications owing to library version mismatch.
As for hyperparameters, we followed the previous work setting.
We confirmed that the DeBERTa model can achieve 0.867 in accuracy despite the smaller training dataset size.

\section{Instruction-Following Ability about length control}
\label{appendix:length_constraint}
Some LLMs prefer to generate longer text.
We compared the instructions for LLMs with and without length constraints.
Table \ref{table:length_constraint} shows the output length difference on instructions across models.
We observe that the FLAN-T5 variants are insensitive to instructions in terms of length.
While other LLMs tend to generate longer text in vanilla instructions, length constraints instructions can shorten the length, the output lengths are proportional to the specified length. 
However, they cannot follow the exact length of words. 
This aligns with the results of prior work~\cite{zhou2023instructionfollowing}. 
\begin{table}[t!]
\centering
\small
\begin{tabular}{@{}llp{3cm}@{}}
\toprule
      & For MRC training & For instruction-following evaluation   \\ \midrule
train & 1690      &  N/A \\
valid & 240      &  N/A \\
test  & 338      &  333\\ \bottomrule
\end{tabular}
\caption{The number of datasets for the MRC model training and instruction-following evaluation.}
\label{table:dataset_size}
\end{table}

\begin{table}[t!]
\small
\centering
\begin{tabular}{@{}lrrr@{}}
\toprule
              & \multicolumn{3}{c}{length constraint} \\ 
model         & w/o    & w/ 15 words   & w/ 10 words  \\\midrule
FLAN-T5-base  & 7.8    & 7.8           & 7.8          \\
FLAN-T5-large & 7.8    & 7.8           & 7.8          \\
FLAN-T5-small & 9.8    & 10.2          & 10.3         \\
FLAN-T5-xl    & 8.2    & 8.9           & 8.9          \\
FLAN-T5-xxl   & 8.0    & 7.8           & 7.8          \\
Llama2-7B     & 38.4   & 16.6          & 13.2         \\
Llama3-8B     & 37.7   & 16.4          & 11.5         \\
Mistral-7B    & 31.8   & 23.8          & 17.3         \\
GPT-3.5       & 25.1   & 18.4          & 13.1         \\\midrule
Oracle        & \multicolumn{3}{c}{15.1}              \\ \bottomrule
\end{tabular}
\caption{The output length difference between instruction w/o and w/ length constraints. In the w/ length constraints setting, we varied the output length as 15 words or 10 words.}
\label{table:length_constraint}

\end{table}

\section{Other Automatic Metrics Confirmation}
\paragraph{Instruction-Following Ability Evaluation using other models.}\label{appendix:human_eval_comaprison}
We compared the models rather than the MRC model using DeBERTa.
We used the BERT model\footnote{\url{https://huggingface.co/google-bert/bert-base-uncased}} as next sentence prediction~(NSP), and GPT-3.5 and GPT-4 as LLMs evaluators for evaluating instruction-following ability on story-ending generation task.

Table~\ref{table:compare_human_eval} shows the result of human evaluation analyzed by evaluation model prediction
as in \S\ref{sec:human_eval}.
This indicates the NSP doesn't have enough ability to distinguish instruction-following due to its objective, and GPT-4 can archive. 
However, the calling API of GPT-4 requires high monetary cost and irregularity of the models' update frequency hinders reproducing evaluation results.
Furthermore, its evaluation has been known to have some biases and to be blackbox, which requires further analysis.

\begin{table}[t]
\centering
\small
\begin{tabular}{@{}llll@{}}
\toprule
model prediction&\multicolumn{3}{c}{Human Evaluation Perspectives} \\
     & Fluency & Coherence & \begin{tabular}{l}
     Instruction\\Following\end{tabular}\\\midrule[0.4mm]
\multicolumn{4}{c}{BERT NSP}          \\ 
Follow     & 4.40  & 4.16 & 3.80   \\
Not Follow & 4.58 & 4.08 & 3.55  \\\hdashline
$\Delta$    & 0.18 & 0.08 & 0.25  \\\bottomrule[0.4mm]
\multicolumn{4}{c}{GPT-3.5}       \\
Follow     & 4.46 & 4.26 & 3.81  \\
Not Follow & 4.66 & 3.71 & 3.12  \\\hdashline
$\Delta$   & 0.2  & 0.55 & 0.69  \\\bottomrule[0.4mm]
\multicolumn{4}{c}{GPT-4}        \\
Follow     & 4.43 & 4.13 & 4.05  \\
Not Follow & 4.68 & 4.06 & 2.88 \\\hdashline
$\Delta$   & 0.25 & 0.07 & 1.17 \\ \bottomrule[0.4mm]
\end{tabular}
\caption{Comparison of Automatic Evaluation on Instruction-Following in terms of Human Evaluation}
\label{table:compare_human_eval}
\end{table}

\paragraph{Dissimilarity using another model.}
When calculating dissimilarity, selecting the model for calculating sentence similarity has several alternatives.
We have confirmed another model\footnote{\url{https://huggingface.co/sentence-transformers/all-MiniLM-L6-v2}} which is famous in Huggingface.
Table~\ref{table:result_dissimilarity_othermodel} shows the dissimilarity result. Its tendency is similar to the dissimilarity calculated by LaBSE.
\begin{table}[t!]
\centering
\small
\begin{tabular}{@{}ll@{}}
\toprule
model                    &   Dissimilarity↑  \\ \midrule
FLAN-T5-small                & 0.387              \\
FLAN-T5-base                  & 0.527            \\
FLAN-T5-large                      & 0.568              \\
FLAN-T5-xl                          & 0.500              \\
FLAN-T5-xxl                              &   0.533              \\
Mistral-7B                          & 0.459              \\
Llama2-7B                &  0.509             \\
Llama3-8B             & 0.600               \\
GPT-3.5              & 0.563              \\ \midrule
Oracle Endings          & 0.631          \\ \bottomrule   
\end{tabular}
    \caption{Dissimilarity using another model}
    \label{table:result_dissimilarity_othermodel}
\end{table}

\begin{table*}[t!]
\centering
\small
\begin{tabular}{@{}llp{12cm}@{}}
\toprule
  & Context     & Kate made a new friend named Jan at the library. They walked to the corner store to buy snacks. Jan bought a can of soda as well. Jan snorted   when she laughed and soda came out of her nose. \\ \midrule
Follow & Instruction 1      & Which is the most likely scenario to happen if the joke was funny to both of the girls?                                                                                                         \\\midrule
× & FLAN-T5-xxl & Jan would have to clean up the soda.                                                                                                                                                              \\
\checkmark & Mistral-7B  & They both laughed heartily, with soda spraying from Jan's nose.                                                                                                                                   \\
\checkmark & Llama3-8B   & Kate laughed just as loudly, soda spraying from her nose too.                                                                                                 \\
\checkmark & GPT-3.5     & They laughed even harder, causing both of them to spill their snacks.                                                                                                                             \\
\checkmark & oracle      & They both laughed as soda spilled out of her nose.                                                                                                                                                \\\midrule
Follow  & Instruction 2      & If Jan was really self conscious about herself, what most likely happened?                                                                                                                        \\\midrule
× & FLAN-T5-xxl & Jan would have snorted less.                                                                                                                                                                      \\
\checkmark & Mistral-7B  & Jan's self-consciousness led her to hide her embarrassing soda-nose laugh from Kate.                                                                                                              \\
\checkmark & Llama3-8B   & Jan quickly apologized and tried to discreetly wipe her nose.                                                                                                                                     \\
\checkmark & GPT-3.5     & Jan felt embarrassed and immediately covered her face with her hands.                                                                                                                             \\
\checkmark & oracle      & Jan became so embarrassed that her whole face turned red.                                                                                                                                         \\ \bottomrule
\end{tabular}
\caption{\textbf{Generation Examples}: The generated endings across some settings on the same context and the different conditions. \textit{Follow} column means whether the label, predicted by the MRC model, is \textit{Follow} or not. Indeed FLAN-T5-xxl's generations seem not to follow the instructions.
Furthermore, even if the endings follow the instructions, there are diverse endings in terms of style and content.}
\label{table:examples}
\end{table*}

\section{Generation Examples}\label{sec:result_detail}
We show the story-endings generated by the instructions across the models, with labels predicted by the MRC model.
Even within the same contexts and instructions, several possible stories to write exist.

\section{Evaluation Interface}\label{sec:evaluation_ui}
\begin{figure*}[t]
    \centering
    \includegraphics[width=\linewidth]{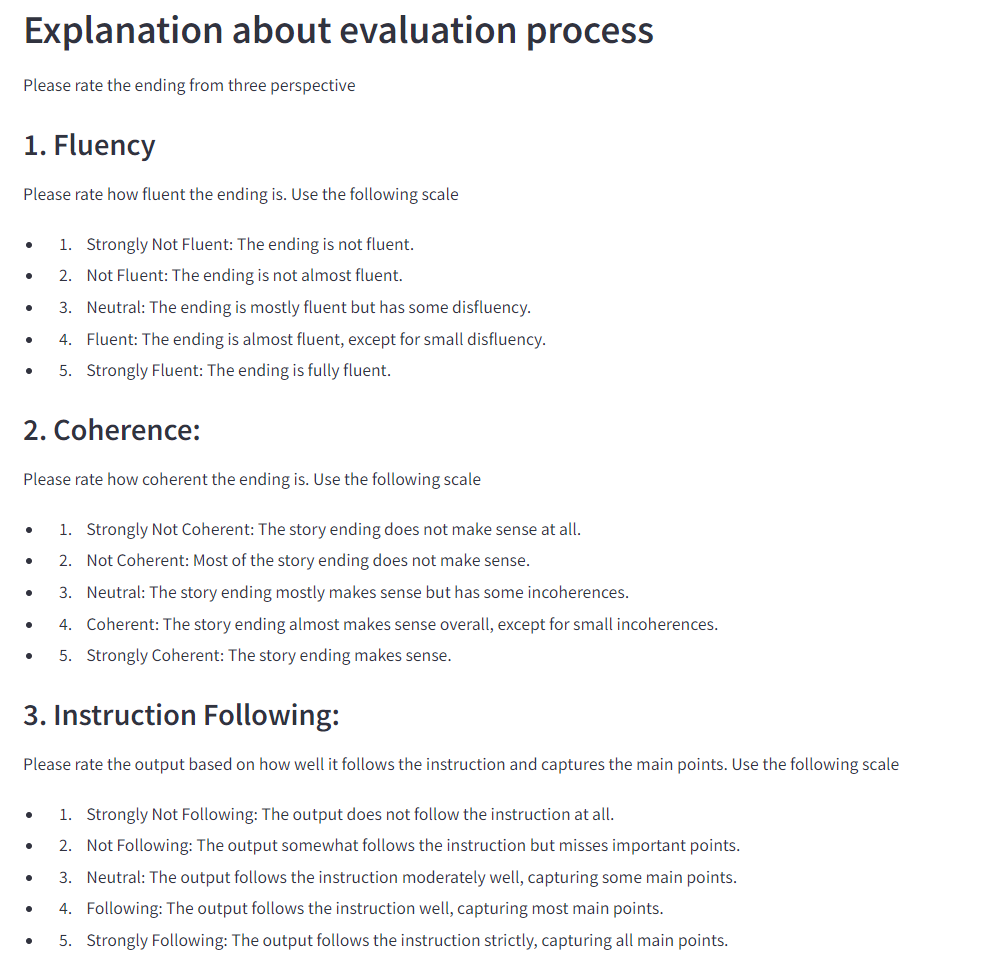}
    \caption{Instruction for Evaluator.}
    \label{fig:evaluation_instruction}
\end{figure*}

\begin{figure*}[t]
    \centering
    \includegraphics[width=1.0\linewidth]{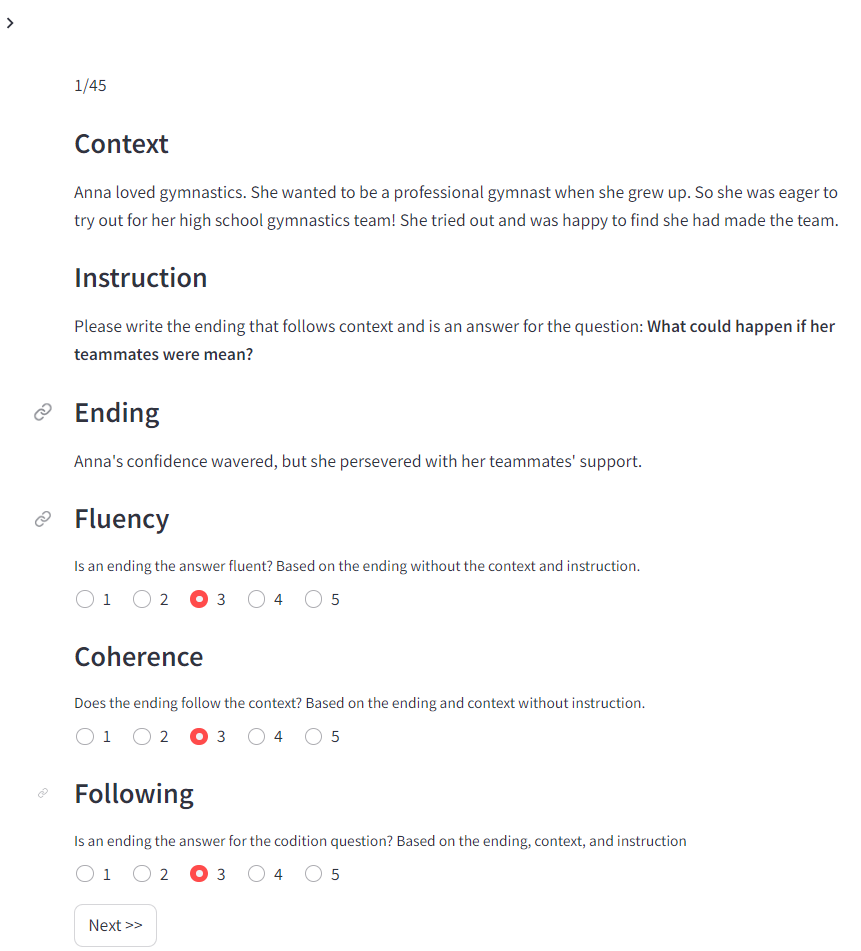}
    \caption{Evaluation Page.}
    \label{fig:evaluation_text}
\end{figure*}

The evaluation interface was implemented using streamlit library\footnote{\url{https://streamlit.io/}}.
Details of the evaluation interface are shown in Figure \ref{fig:evaluation_instruction} and \ref{fig:evaluation_text}. 


\section{Comparison Models Detail} \label{comparison_model_info}
We downloaded open-source models from Huggingface\footnote{\url{https://huggingface.co/models}}.
We also used the GPT-3.5 API via OpenAI API\footnote{\url{https://openai.com/api/}}.
Table \ref{table:comparison_models} shows the checkpoints and licenses of the comparison models.

\begin{table*}[t]
\small
\centering
\begin{tabular}{@{}lll@{}} \toprule
model         & checkpoint                                          &license            \\\midrule
FLAN-T5-small & \url{https://huggingface.co/google/flan-t5-small}    & Apache 2.0            \\
FLAN-T5-base  & \url{https://huggingface.co/google/flan-t5-base}     & Apache 2.0             \\
FLAN-T5-large & \url{https://huggingface.co/google/flan-t5-large}    & Apache 2.0                                                 \\
FLAN-T5-xl    & \url{https://huggingface.co/google/flan-t5-xl}       & Apache 2.0                                            \\
FLAN-T5-xxl   & \url{https://huggingface.co/google/flan-t5-xxl}      & Apache 2.0                                               \\
Llama2-7B    & \url{https://huggingface.co/meta-llama/Llama-2-7b-chat-hf} & custom license\\
Llama3-8B    & \url{https://huggingface.co/meta-llama/Meta-Llama-3-8B-Instruct} & custom license\\
Mistral-7B    & \url{https://huggingface.co/mistralai/Mistral-7B-Instruct-v0.2}& Apache 2.0   \\
GPT-3.5    &  gpt-3.5-turbo-0613&proprietary\\\bottomrule

\end{tabular}
\caption{Comparison Model Detail}
\label{table:comparison_models}

\end{table*}

\end{document}